\documentclass{article}

\usepackage{microtype}
\usepackage[dvipdfmx]{graphicx}
\usepackage{booktabs} 

\usepackage{hyperref}

\usepackage[accepted]{icml2020}

\usepackage{mathrsfs}
\usepackage{times}
\usepackage{soul}
\usepackage{url}
\usepackage[utf8]{inputenc}
\usepackage[small]{caption}
\usepackage[dvipdfmx]{graphicx}
\usepackage{amsmath}
\usepackage{amsfonts}
\usepackage{amsthm}

\usepackage{multirow}

\usepackage{amsmath,amssymb,amsfonts,amsthm,enumerate}
\usepackage{bussproofs}
\usepackage[subrefformat=parens]{subcaption}
\usepackage{scalerel}

\theoremstyle{definition}
\newtheorem{theorem}   {Theorem}  [section]

\newtheorem{defi} [theorem] {Definition}
\newtheorem{example}    [theorem] {Example}

\numberwithin{equation}{section}

\newcommand{\softmax}{\mathrm{softmax}}
\newcommand{\reshape}{\mathrm{reshape}}

\icmltitlerunning{Learning Directly from Grammar Compressed Text}

\begin{document}

\twocolumn[
\icmltitle{Learning Directly from Grammar Compressed Text}



\icmlsetsymbol{equal}{*}

\begin{icmlauthorlist}
\icmlauthor{Yoichi Sasaki}{equal,to}
\icmlauthor{Kosuke Akimoto}{equal,to}
\icmlauthor{Takanori Maehara}{goo}
\end{icmlauthorlist}

\icmlaffiliation{to}{NEC Corporation, Kananagawa, Japan}
\icmlaffiliation{goo}{RIKEN Center for Advanced Intelligence Project (RIKEN~AIP), Nihonbashi, Tokyo, Japan}

\icmlcorrespondingauthor{Yoichi Sasaki}{yoichisasaki@nec.com}
\icmlcorrespondingauthor{Kosuke Akimoto}{kosuke\_a@nec.com}


\vskip 0.3in
]



\printAffiliationsAndNotice{\icmlEqualContribution} 
\renewcommand{\algorithmiccomment}{$\ \ \ \ \triangleright$}

\begin{abstract}
Neural networks using numerous text data have been successfully applied to a variety of tasks.
While massive text data is usually compressed using techniques such as grammar compression, almost all of the previous machine learning methods assume already decompressed sequence data as their input.
In this paper, we propose a method to directly apply neural sequence models to text data compressed with grammar compression algorithms without decompression.
To encode the unique symbols that appear in compression rules, we introduce composer modules to incrementally encode the symbols into vector representations.
Through experiments on real datasets, we empirically showed that the proposal model can achieve both memory and computational efficiency while maintaining moderate performance.
\end{abstract}
\section{Introduction}
\label{sec:introduction}

Machine learning methods have been successfully applied to huge text data such as large text corpora~\cite{devlin-etal-2019-bert,radford2019language} and multiple genome sequences~\cite{phaml2005qualitatively}.
In practice, large text data are usually stored in a compressed format. 
For example, genome sequences are treated and maintained in a compression format for each individual, and news items are compressed into a certain time period.
Standard machine-learning models can only take an original (i.e., non-compressed) text as an input, and thus, a naive method for these compressed text data involves a decompression step.
However, decompression can be expensive because its running time depends on the size of the original uncompressed text, which can be much longer than its compressed text.
Also, since large data makes machine learning time- and memory-consuming, it is undesirable to decompress and inflate size of data especially if the compression ratio of the data is high.

In this paper, we propose an efficient machine learning method that does not involve any decompression process and directly conducts inference and learning on compressed texts.
More specifically, the proposed method directly applies neural sequence models to text data that are compressed with \emph{grammar compression} algorithms.

Grammar compression algorithms are commonly used to compress text data since they allow the original text to be perfectly reconstructed from the compressed text, i.e., they are lossless compression, and it is known that they can effectively compress text in general~\cite{DBLP:conf/dcc/LarssonM99,DBLP:conf/cpm/GotoBIT15}.
Technically, a grammar compression algorithm encodes an original text into a compressed sequence and a set of replacement rules of context-free grammar.
A compressed sequence consists of \emph{terminals}, which are the alphabets appeared in the original text, and \emph{non-terminals}, which are new alphabets generated by a replacement rules;
see Section~\ref{sec:preliminary} for the precise definition.

Since replacement rules of a compressed text may contain information of sensitive sub-parts of the original uncompressed text, it is not desirable to compress texts using a shared single set of replacement rules constructed on a whole corpus in applications that require privacy such as genome classification.
Grammar compression algorithms can be also applied to such applications because they can construct unique replacement rules and non-terminals for each text.
Thus, in this paper, we focus on compressed texts which are independently compressed by grammar compression algorithms.
However, independently compressed text has different rules and non-terminals, and thus, we cannot naively replace non-terminals by embedding vectors, making it difficult to directly apply neural sequence models to grammar compressed text.
To circumvent this difficulty, the proposed method uses a trainable composer module to incrementally compute vector representations of non-terminals from trainable embeddings of terminals based on replacement rules, making it possible to represent a compressed text by a sequence of vectors which can be input to existing neural sequence models.

Since length of the sequence of vector is same as that of the compressed text which is shorter than the uncompressed text, it is expected that the proposed method can reduce memory and computational cost of neural sequence models.
Also, the proposed method employs a clustering algorithm of non-terminals to parallelize the computation of vector representations of non-terminals to improve efficiency.
We empirically show efficiency of the proposed method through experiments on real datasets\footnote{Our source code is available at \url{https://github.com/aurtg/learning-on-grammar-compression}.}.

Our contributions are summarized as follows:
\begin{itemize}
    \item We propose a method to apply neural sequence models to grammar compressed text data without decompression.
    Also, we propose a clustering algorithm of non-terminals to improve efficiency of the proposed method by parallel computation.
    \item Through experiments on real datasets, we empirically showed that the proposed model can achieve both memory and computational efficiency while maintaining moderate performance.
\end{itemize}

\section{Related Work}

\subsubsection*{Algorithms on compressed text and matrix}

Text mining or matching on compressed text data has been extensively studied for many years~\cite{DBLP:conf/cpm/BilleFG07,DBLP:journals/jda/Navarro03,DBLP:conf/dcc/GasieniecKPS05}.

To accelerate basic matrix operation, compression-based methods, which is aim to in-memory processing, have been developed~\cite{DBLP:journals/pvldb/Rendle13,DBLP:conf/kdd/TabeiSYP16,DBLP:journals/jip/OyamadaLINAK18}.
After the each column of the matrix is compressed, for matrix operation, the methods conduct access to elements of the matrix without decompression.

\subsubsection*{Learning directly from compressed image}

While there is no existing method that learns from compressed text, there exists a method that learns directly from compressed image, which represented as JPEG format~\cite{DBLP:conf/nips/GueguenSKLY18,ulicny2017using}.
JPEG image data from the original RGB data is encoded in three steps: first, the RGB image is converted into the YCbCr color space and the chroma channels are downsampled, next, the channels are projected through the discrete cosine transform (DCT) and quantized, and finally, the quantized coefficients are losslessly compressed by using Huffman codes (e.g., RGB $\rightarrow$ YCbCr $\rightarrow$ Quantized DCT $\rightarrow$ Huffman codes steps).
\cite{DBLP:conf/nips/GueguenSKLY18} run only the first step of decoding and then feed the DCT coefficients directly into a neural network.
\cite{DBLP:conf/nips/GueguenSKLY18} extend \cite{ulicny2017using} to the full early JPEG stack, to which allows for much deeper networks and for training on a much larger dataset and more difficult tasks.

\subsubsection*{Memory Efficient Neural Sequence Models}

There are a few lines of existing research which are investigating methods to make neural sequence models memory efficient.

Following idea of trading computation time and memory consumption during back-propagation \cite{dauvergne2006data,chen2016training,DBLP:conf/nips/VaswaniSPUJGKP17},
in \cite{DBLP:conf/nips/GruslysMDLG16}, the authors proposed efficient approach which reduce memory consumption of the backpropagation when training recurrent neural networks with additional computational costs.

More recently, \cite{kitaev2020reformer} proposed various techniques including approximate attention computation to reduce memory usage of Transformers from $O(n^2)$ to $O(n\log n)$.

\section{Preliminaries}
\label{sec:preliminary}

\subsection{Notations}

For a finite set $C$, the cardinality of $C$ is denoted by $|C|$.
Let $\Sigma$ be a finite set of symbols.
The set of all possible sequences of $\Sigma$ is denoted by $\Sigma^{\ast}$.
For a sequence $S$, the length of $S$ is denoted by $|S|$.
For $i = 1, \dots, |S|$, the $i$-th symbol of $S$ is denoted by $S[i]$.
The contiguous subsequence that begins at position $i$ and ends at position $j$ is denoted by $S[i..j]$.

\subsection{Grammar Compression}
\label{subsec_grammar_comp}
A \textit{context free grammar (CFG)} is a collection of rules of the form: $A \rightarrow a_1 a_2 \dots a_k$.
The symbols appearing on the left-hand side of the rules are called \textit{non-terminals}, and the symbols appearing on the input sequence called \emph{terminals}.
Note that both terminals and non-terminals appear on the right-hand side of the rules.

Given the original sequence $S$, a \textit{grammar compression method} constructs a set of non-terminals $V$, a compressed sequence $C\in (\Sigma\cup V)^*$, and a set of restricted CFG rules $R$, from which we can uniquely derive the original sequence $S$.
We define a \textit{compression tuple} of $S$ as a tuple $S_{comp}=(V, R, C)$.

In this paper, we focus on grammar compression methods whose each replacement rule in $R$ contains only two symbols in its right-hand side unless otherwise specified.
Also, without loss of generality, we assume that both right-hand symbols $a_1^{(i)}, a_2^{(i)}$ in $i$-th replacement rule $A^{(i)}\rightarrow a_1^{(i)} a_2^{(i)}$ in $R$ are terminal symbols or appear in right-hand side of previous rules, i.e., $a_1^{(i)}, a_2^{(i)}\in (\Sigma \cup \{ A^{(1)},...,A^{(i-1)}) \}$.

Below, we introduce three commonly-used grammar compression algorithms.

\subsubsection{Re-Pair Algorithm}
The \emph{recursive pairing compression algorithm (Re-Pair)} is a grammar compression algorithm proposed by Larsson and Moffat~\yrcite{DBLP:conf/dcc/LarssonM99}.
This algorithm repeatedly finds a most frequently appearing bi-gram and replaces it with a new non-terminal symbol until the replacement does not reduce the representation cost (i.e., the sum of the length of the compressed sequence and the sizes of the rules).
Here, each replacement rule is inserted into a rule dictionary.

\begin{example}
Suppose that we have the following sequence:
\begin{equation}
\label{seq}
    S = aababcababcabcd.
\end{equation}
Then, Re-Pair gives the following compressed sequence $C$ and the rule dictionary $R$:
\begin{equation}
\begin{split}
    C&=aABABBd, \\ 
    R&=\{A\rightarrow ab, B\rightarrow Ac\}.
\end{split}
\end{equation}
Re-Pair first replaces $ab$ with a new non-terminal symbol $A$, as $ab$ is the most frequently appearing pair.
Then it recounts the frequency of pairs in the replaced sequence $aAAcAAcAcd$.
It replaces $Ac$ with $B$ and stops this procedure when it can no longer reduce the representation cost.
\end{example}

\subsubsection{LZ78 and LZD Algorithm}
The \emph{LZ78 algorithm} was proposed by Lempel and Jacob Ziv~\yrcite{DBLP:journals/tit/ZivL78}.
This algorithm works by constructing a dictionary of parts, which we call \textit{phrases}, that have appeared in the sequence.
Assume we are compressing a sequence $S[1..n]$ and have already processed $S[1..i-1]$ into $r$ phrases $P_0 P_1 ..P_{r-1}$, where phrase $P_0$ represents an empty sequence for convenience.
Then, to compute $P_r$, we find the longest prefix $S[i..j-1]$ (with $j-1 < n$) that is equal to some $P_q$, with $q < r$.
Here we define $P_r = P_q T[j]$, which is represented as the pair $(q, T[j])$.

\begin{example}
Suppose that we have the string (\ref{seq}).
LZ78 starts with the shortest phrase on the left that we haven't seen before.
This will always be a single symbol; in this case, $P_1 = a$.
It now takes the next phrase we haven't seen.
We have already seen $a$, so it takes $P_2 = ab$.
Continuing, finally, we get the following sequence of phrases:
\begin{equation}
    a|ab|abc|aba|b|c|abcd
\end{equation}
and obtains the rule dictionary $R$ and compressed sequence $C$ as
\begin{equation}
\label{LZ78example}
\begin{split}
    C=&P_0P_1P_2P_3P_4P_5P_6P_7 \\ 
    R=&\{P_0\rightarrow \ ,P_1\rightarrow P_0a, P_2\rightarrow P_1b, P_3\rightarrow P_2c,\\ &\ P_4\rightarrow P_2a, P_5\rightarrow P_0b, P_6\rightarrow P_0c, P_7\rightarrow P_3d \}.
\end{split}
\end{equation}

In reality, these can be efficiently represented by\footnote{These can also be represented by the sequence of 2-tuples corresponding to the right-hand symbols of the rules in (\ref{LZ78example}) such as $(P_0, a)(P_1, b)(P_2, c)(P_2, a)(P_0, b)(P_0, c)(P_3, d)$.} 
\begin{equation}
\begin{split}
    C&=aP_2P_3P_4bcP_7 \\ 
    R&=\{P_2\rightarrow ab, P_3\rightarrow P_2c, P_4\rightarrow P_2a, P_7\rightarrow P_3d \}.
\end{split}
\end{equation}
\end{example}

The LZ double algorithm (LZD)~\cite{DBLP:conf/cpm/GotoBIT15} is an extension of LZ78.
\cite{DBLP:conf/cpm/GotoBIT15} simply change the definition of a phrase $P_i$ to the pair of the longest previously occurring phrase $P_{j_1}$ and the longest previously occurring phrase $P_{j_2}$ that also appears at position $|P_1|+\dots +|P_{i-1}|+|P_{j_1}|+1$.

\begin{example}
Suppose that we have the sequence (\ref{seq}).
The LZD phrase of the sequence (\ref{seq}) is
\begin{equation}
    a|ab|abc|ababc|abcd
\end{equation}
and can be represented by
\begin{equation}
\begin{split}
    C&=aP_2P_3P_4P_5 \\ 
    R&=\{P_2\rightarrow ab, P_3\rightarrow P_2c, P_4\rightarrow P_{2}P_{3}, P_5\rightarrow P_3d \}.
\end{split}
\end{equation}
\end{example}

\subsection{Neural Sequence Models}
\label{subsec_neural_sequence_models}

Various neural network architectures have been applied to sequence data, including recurrent neural network (RNN) \cite{DBLP:journals/cogsci/Elman90}, long short-term memory (LSTM) \cite{DBLP:journals/neco/HochreiterS97}, gated recurrent unit (GRU) \cite{cho-etal-2014-properties}, convolutional neural network (CNN) \cite{DBLP:conf/icml/LeM14}, and Transformers \cite{DBLP:conf/nips/VaswaniSPUJGKP17}.

Mathematically, these models can be represented as follows:
\begin{equation}
    ( \boldsymbol{y}_1,...,\boldsymbol{y}_n ) = f(( \boldsymbol{x}_1,...,\boldsymbol{x}_n )),
\end{equation}
where $n$ is the length of the input sequence, and $\boldsymbol{x}_t\in\mathbb{R}^{d_i}$ and $\boldsymbol{y}_t\in\mathbb{R}^{d_o}$ are input and output vector representations of the $t$-th entry of the input sequence, respectively.
In classification and regression tasks, the final prediction for each input sequence is generally computed from output vector representations $\boldsymbol{y}_1, ... , \boldsymbol{y}_n$.

While neural sequence models have been successfully applied to various tasks involving sequence data, they require at least $O(n)$ time, where $n$ is the length of the input sequence \cite{DBLP:conf/nips/VaswaniSPUJGKP17}.
\section{Proposed Method}

\subsection{Task Definitions}

We focus on sequence classification tasks in which each input text is compressed by a grammar compression algorithm\footnote{Note that the proposed method can easily be extended to regression tasks.}, i.e., $i$-th input $S_{comp,i}$ is represented by its compression tuple $S_{comp,i}=(V_i, R_i, C_i)$, where $V_i$ is non-terminals in $S_{comp,i}$, $R_i$ is a rule dictionary corresponding to $S_{comp,i}$, and $C_i$ is a compressed sequence.
Note that we use different CFGs for each sequence.
We call this setting \emph{individual compression}.
This setting is suitable for applications mentioned in Section~\ref{sec:introduction}.

\subsection{Encoding Non-Terminals}
\label{subsec_encoding_nonterminals}

We can assign a trainable parameter vector $\boldsymbol{x}_c$ for each terminal symbol $c \in \Sigma$ because the terminal symbols are common for all input sequences.
However, we cannot assign a trainable vector for non-terminal symbols in $V_i$ because the non-terminal symbols are generated for each input sequence;
therefore, they are not shared across different sequences.

Here, we propose a method to obtain vector representations of non-terminals by composing vector representations of the terminals.

\subsubsection*{Composer Module}
For the $k$-th replacement rule $A^{(k)}\rightarrow a_1^{(k)} a_2^{(k)}$ in $R_i$ and vector representations of the symbols in the right-hand, $\boldsymbol{x}_{a_1^{(k)}}, \boldsymbol{x}_{a_2^{(k)}}$, we compute a vector representation of the left-hand symbol $\boldsymbol{x}_{A^{(k)}}$ by using \textit{composer module} $M$ as follows:
\begin{align}
\boldsymbol{x}_{A^{(k)}}=M(\boldsymbol{x}_{a_1^{(k)}}, \boldsymbol{x}_{a_2^{(k)}}; \theta).
\end{align}
Here, $M$ is a function that has a trainable parameter $\theta$.
In this paper, we test two composer modules, \textit{multi-layer perceptron (MLP) composer} $M_\text{MLP}$ and \textit{dual-GRU composer} $M_\text{GRU}$\footnote{\cite{DBLP:conf/acl/TaiSM15} proposed Binary Tree-LSTMs, which have similar design to our dual-GRU composer modules, and applied it to dependency and constituency tree. We did not test Binary Tree-LSTMs since it did not perform better than our Dual-GRU composer in preliminary experiments.
}.

\textbf{MLP:} As a baseline, we use the multi-layer perceptron $f_{MLP}(\cdot; \theta): \mathbb{R}^{2d}\rightarrow \mathbb{R}^d$ as a composer module:
\begin{equation}
    M_\text{MLP}(\boldsymbol{x}_1, \boldsymbol{x}_2; \theta)=f_{\text{MLP}}([\boldsymbol{x}_1;\boldsymbol{x}_2]; \theta).
\end{equation}
Here, $[\cdot ; \cdot]$ denotes the vector concatenation.

\textbf{Dual-GRU:} 
Because we recursively apply composer modules to encode non-terminals, the MLP composer may cause exploding and/or vanishing gradient problems.
To alleviate this problem, we use a slightly modified version of GRU cell ~\cite{cho-etal-2014-properties} as a composer module:

\begin{equation}
    M(\boldsymbol{x}_1, \boldsymbol{x}_2; \theta) = \boldsymbol{z}_1 \circ \boldsymbol{x}_1 + \boldsymbol{z}_2 \circ \boldsymbol{x}_2 + \boldsymbol{z}_i \circ \boldsymbol{i},
\end{equation}
where $\circ$ is the element-wise product, and
$\boldsymbol{z}_1$, $\boldsymbol{z}_2$, $\boldsymbol{z}_i$, and $\boldsymbol{i}$ are defined by the following formulas:
\begin{align}
    [\boldsymbol{z}_1, \boldsymbol{z}_2, \boldsymbol{z}_i] =& \softmax(\reshape(W_z[\boldsymbol{x}_1;\boldsymbol{x}_2] + \boldsymbol{b}_z)), \nonumber \\
    \boldsymbol{r} =& \sigma(W_r[\boldsymbol{x}_1;\boldsymbol{x}_2] + \boldsymbol{b}_r), \nonumber \\
    \boldsymbol{i} =& \sigma(W_i(\boldsymbol{r}\circ [\boldsymbol{x}_1;\boldsymbol{x}_2]) + \boldsymbol{b}_i),
\end{align}
where $\softmax: \mathbb{R}^{3\times d}\rightarrow \mathbb{R}^d\times \mathbb{R}^d \times\mathbb{R}^d$ denotes a function that applies the softmax function to each row of the input matrix and outputs each column,
$\reshape: \mathbb{R}^{3d} \rightarrow \mathbb{R}^{3\times d}$ is the function that converts a 1D vector into a 2D matrix of a suitable shape, and $\sigma$ denotes the element-wise sigmoid function.
$W_z\in\mathbb{R}^{3d\times 2d}, \boldsymbol{b}_z\in\mathbb{R}^{3d}, W_r\in\mathbb{R}^{2d\times 2d}, \boldsymbol{b}_r\in\mathbb{R}^{2d}, W_i\in\mathbb{R}^{d\times 2d}, $ and $\boldsymbol{b}_i\in\mathbb{R}^d$ are trainable parameters $\theta$ of the composer.

\subsubsection*{Non-Terminal Encoder}

As described in Section~\ref{subsec_grammar_comp}, right-hand symbols in each rule are either a terminal symbol or a left-hand symbol in a previous rule.
Thus, given the vector representations of terminal symbols, we can compute the vector representations of all non-terminals by sequentially applying a composer module to those vector representations of the right-hand symbols of each rule.

\subsection{Clustering Non-terminals for Parallel Computation}
\label{subsec:clustering_non_terminals}
To take advantage of the massively parallel computing capabilities of GPUs, we propose a clustering algorithm for non-terminals to compute the vectors of non-terminal symbols in parallel.

First, we introduce a definition for a directed task graph~\cite{S.Vivek1989}.
\begin{defi}
A \textit{directed task graph} is denoted as $G=(V, E, \mathscr{T}, \mathscr{C})$, where $V$ is a set of task nodes, $E$ is a set of edges, $\mathscr{T}$ is the set of node computation costs, and $\mathscr{C}$ is the set of edge communication costs. 
$t_i \in \mathscr{T}$ is the execution time of node $v_i \in V$
and $c_{i,j} \in \mathscr{C}$ is the communication cost incurred along the edge $e_{i,j}=(n_i, n_j)\in E$, which is zero if both nodes are mapped on the same processor. 
\end{defi}

\textit{Clustering} is a mapping of the task nodes in the graph into clusters.
A cluster is a set of tasks that will be executed on the same processor.
The clustering problem is known to be NP-complete for a general task graph and for several cost functions~\cite{DBLP:journals/siamcomp/PapadimitriouY90,S.Vivek1989}.
Even if the cost function is the minimization of parallel time with an unbounded number of processors, this problem is NP-hard~\cite{DBLP:journals/siamcomp/PapadimitriouY90,S.Vivek1989}.

To overcome this complexity, we consider a simple task graph $G=(V, E, \mathscr{T}, \mathscr{C})$, where $V$ indicates tasks that compute the representations for non-terminals, $E$ indicates the inverse dependencies corresponding to rules, all the values in $\mathscr{T}$ are equal, and all the values in $\mathscr{C}$ are zero\footnote{Here, we assumes that computational cost of a composer module is equal with fixed size of input vectors and that we can ignore time of memory access and communication between CPU and GPU.}.
Furthermore, computational cost of a composer module can be assumed to be equal with fixed size of input vectors.

\begin{algorithm}[t]
	\caption{Clustering Algorithm}
	\label{clustering}
	\begin{algorithmic}[1]
	    \STATE {\textbf{procedure} \textsc{Main}$(G)$}
		\STATE {\textbf{Input:} A directed task graph $G=(V,E)$}
		\STATE {\textbf{Output:} Lists of executable nodes in parallel for each depth $L_0,..,L_d$, Clusters of input graph $C_1,..,C_p$} 
		\STATE {\textsc{ParallelNode}$(G,0)$;} \COMMENT{Get $L_1,..,L_d$}
		\FOR{$i=0,..,d$}
		\STATE {Assign each node in $L_i$ to a different cluster in order from $C_1$}
		\ENDFOR
		\newline
		\STATE  {\textbf{procedure} \textsc{ParallelNode}$(G=(V,E), i)$}
        \STATE  {$E'\leftarrow E$} \COMMENT{Initialization}
        \IF     {$E=\emptyset$}
        \STATE  {\textbf{return}}
        \ELSE
		\FOR {$x\in V'=\{v|v\in V$, indegree of $v$ is 0$\}$}
		\STATE  {$L_i\leftarrow L_i + \{x\}$}
		\STATE  {$E_{out}(x)=\{(x,y)|y \in V, (x,y) \in E \}$}
		\STATE  {$E'\leftarrow E'-E_{out}(x)$}
		\ENDFOR
		\STATE  {Report $L_i$ as executable nodes in parallel for depth $i$}
		\STATE  {\textsc{ParallelNode}$(G'=(V,E'), i+1)$} \COMMENT{Recursive call}
        \ENDIF
	\end{algorithmic}
\end{algorithm}

For this task graph, we give the simple clustering algorithm in Algorithm~\ref{clustering}.
In this algorithm, the subprocedure \textsc{ParallelNode} starts with the input task graph $G$ and $i=0$.
Then it adds nodes whose indegree is zero to the list $L_0$ as the executable task nodes in parallel.
Here, for a vertex $x$, the number of edges whose heads end adjacent to $x$ is called the indegree of the vertex $x$.
It recursively calls with $i+1$ and $G'=(V,E')$ where $E'$ is edges excluding edges whose tail nodes are in $L_i$.
Finally, after it calculates all of the lists of executable nodes in parallel, it assigns each node in the list for each depth to a different cluster.

As for the time complexity of this algorithm, it runs in $O(V)$-time, which is exactly equal to the number of rules $R$, since line 4 in Algorithm~\ref{clustering} traverses all edges only once.
Note that, as described in Section~\ref{subsec_grammar_comp}, since we assume each replacement rule contains only two symbols in its right-hand side, the edges in our task graph are twice the number of the nodes.

\subsection{Complexity}
\label{subsec:complexity}

With an input sequence of length $n$, a $l$-layer sequence model requires $O(ln(nd+d^2))$ (Transformer) or $O(lnd^2)$ (CNN, RNN) time to process each sequence, where $d$ is the dimension of vector representations \cite{DBLP:conf/nips/VaswaniSPUJGKP17}.\footnote{We assumes the kernel size of CNN is sufficiently smaller than the dimensions of vector representations $d$.}
On the other hand, with a compressed sequence of length $n'$ and $|R|$ replacement rules, the proposed method requires $O(|R|d^2)$ time to encode all non-terminals with a composer module. Since the proposed method inputs a compressed sequence of length $n'$ to a sequence model, its overall time complexity is either $O((|R|+ln')d^2 + ln'^2d)$ (Transformer) or $O((|R|+ln')d^2)$ (CNN, RNN).

With an input sequence of length $n$, a $l$-layer sequence model requires $O(l(n^2+dn))$ (Transformer) or $O(lnd)$ (CNN, RNN) memory space to store intermediate vectors for back propagation. Since the proposed method requires additional $O(|R|d)$ memory space to store intermediate vectors of composer modules and inputs a compressed sequence of length $n'$ to a sequence model, the overall memory complexity during the training phase is either $O((|R|+ln')d + ln'^2)$ (Transformer) or $O((|R|+ln')d)$ (CNN, RNN).

The above complexity analysis shows that the proposed method can effectively reduce computational complexity and memory complexity during the training phase for compressed sequences with high compression ratios (i.e., $|R|+n'< n$).
Also, note that the above complexity analysis do not change with the clustering algorithm described in Section~\ref{subsec:clustering_non_terminals}.

\section{Experiments}

\subsection{Dataset}

We applied the proposed method to two sequence classification tasks: character-level text classification and DNA sequence classification.
For each of the datasets used in the tasks, we compressed sequences using Re-Pair and LZD algorithms.
The average sequence length of the original and compressed sequences is shown in Table \ref{tab:ave_str_len}\footnote{Since the sequence lengths of 10 DNA classification datasets are almost the same, we show their range in Table~\ref{tab:ave_str_len}.}.
Also, Table~\ref{tab:avg_n_rule} shows the average number of replacement rules of Re-Pair and LZD algorithms in each dataset.

\begin{table*}[]
    \caption{Compression results.}
    \label{tab:stat_comp}
    \begin{minipage}[b]{0.6\hsize}
        \subcaption{Sequence length of sequence classification datasets}
        \label{tab:ave_str_len}
        \centering
        \scalebox{0.9}{
        \begin{tabular}{c||c|c|c}
            \multirow{2}{*}{Dataset}  & \multicolumn{3}{c}{\begin{tabular}{c}
                 sequence length\\
                 (average / max)
            \end{tabular}} \\\cline{2-4}
             & (uncompressed) & (Re-Pair) & (LZD) \\ \hline\hline
            Sogou & $2.69\times 10^{3}$ / $2.27\times 10^{5}$ & 480 / $1.60\times 10^{4}$ & 524 / $1.69\times 10^{4}$ \\ \hline
            Yelp P. & 725 / $5.27 \times 10^{3}$ & 303 / $1.76 \times 10^{3}$ & 255 / $1.31 \times 10^{3}$ \\ \hline
            Yelp F. & 732 / $5.63 \times 10^{3}$ & 306 / $1.78 \times 10^{3}$ & 257 / $1.30 \times 10^{3}$ \\ \hline \hline
            DNA & 500 &  \begin{tabular}{c} 170 -- 171 /\\ 193 -- 194\end{tabular} & \begin{tabular}{c} 119 -- 120 / \\ 127 \end{tabular}
        \end{tabular}}
    \end{minipage}
    \begin{minipage}[b]{0.4\hsize}
        \subcaption{Average number of replacement rules $\overline{|R|}_{r}$ and $\overline{|R|}_{L}$ of Re-Pair and LZD algorithms}
        \label{tab:avg_n_rule}
        \centering
        \begin{tabular}{c||c|c}
           Dataset  & $\overline{|R|}_{r}$ & $\overline{|R|}_{L}$ \\ \hline\hline
            Sogou & 232 & 484 \\ \hline
            Yelp.P & 93.1 & 224 \\ \hline
            Yelp.F & 94.1 & 226 \\\hline\hline
            DNA & 45 -- 46 & 114 -- 115\\
        \end{tabular}
    \end{minipage}
\end{table*}

\subsubsection{Character-Level Text Classification}

We conducted experiments on the character-level text classification dataset\footnote{\url{https://github.com/zhangxiangxiao/Crepe}} in \cite{DBLP:conf/nips/ZhangZL15}.
Among the eight datasets released in the paper, we used Sogou News (Sogou), Yelp Review Polarity (Yelp P.), and Yelp Review Full (Yelp F.), as these three have a longer average sequence length than the other five.
The Sogou and Yelp F. datasets have five class categories and the Yelp P. dataset has two.
The alphabet used in our experiments consists of the 70 characters used in \cite{DBLP:conf/nips/ZhangZL15} and one blank space character. We did not distinguish between lower and upper case letters in our experiments.

\subsubsection{DNA Sequence Classification}

We conducted experiments on 10 DNA sequence classification datasets\footnote{\url{http://www.jaist.ac.jp/~tran/nucleosome/members.htm}} in \cite{phaml2005qualitatively} which are constructed using DNA sequences reported in \cite{pokholok2005genome}, each of which contains DNA sequences relating to a specific type of histone protein.
The number of class categories of all dataset is two.
Sequences in the datasets consist of four characters, $\texttt{A}, \texttt{G}, \texttt{C},$ and $ \texttt{T}$, which correspond to four nucleobases of DNA. The length of each sequence is fixed to 500.
CNN has recently been successfully applied to these dataset with domain-specific 3-gram features~\cite{nguyen2016dna}.

\subsection{Method}

\textbf{Baseline:}
As a simple baseline, we applied sequence models to the original decompressed sequences.
To encode the decompressed sequences, we encode each character $c\in\Sigma$ in the sequence by its corresponding trainable parameter vector $\boldsymbol{x}_c\in\mathbb{R}^d$.

\textbf{Proposed Method:} We tested two types of composer modules described in $\S$\ref{subsec_encoding_nonterminals}. As $f_{\text{MLP}}$ in a naive MLP composer, we used a multi-layer perceptron with a single hidden layer. The dimension of the hidden layer is set to $d$. We used ReLU \cite{DBLP:conf/icml/NairH10} and sigmoid activation in the hidden and output layers, respectively\footnote{Using ReLU activation in the output layer caused explosion of the output representations.}.

\textbf{Sequence Model:} For both the baseline and proposed method, we used a bidirectional LSTM (Bi-LSTM) \cite{DBLP:journals/tsp/SchusterP97} as a sequence model. The dimension of the input and hidden representations $d$ is set to 200, and the number of Bi-LSTM layers is set to one. We max-pooled the outputs of Bi-LSTM to compute a single representation of the input sequence. We applied dropout of $p=0.5$ to the max-pooled output of Bi-LSTM in the text classification datasets. Linear transformation was applied to the sequence representation to compute a score for each category label.

We trained the models using Adam optimizer \cite{DBLP:journals/corr/KingmaB14} with a learning rate of $10^{-3}$ for 50 epochs and halved the learning rate every 10 (text) or 20 (DNA) epochs. Also, we linearly increased the learning rate from $0$ to $10^{-3}$ during the first 1000 steps. We set the mini-batch size to 126 (text) or 10 (DNA) and randomly sampled $100000$ training sequences every epoch for the text classification datasets. We randomly sampled $20\%$ of the training data as a held-out development set and report the performance of the best performing model on the development set. To decide the hyper-parameter settings described above, we empirically tuned the learning rate, dropout rate, weight decay ratio, and frequency of learning rate decay on the basis of the performance of the baseline method on a development set.

\subsection{Results}

\subsubsection*{Classification Performance}
\label{subsubsec_classification_performance}

\begin{figure*}[]
    \begin{minipage}[b]{0.499\linewidth}
        \centering
        \includegraphics[width=0.8\columnwidth]{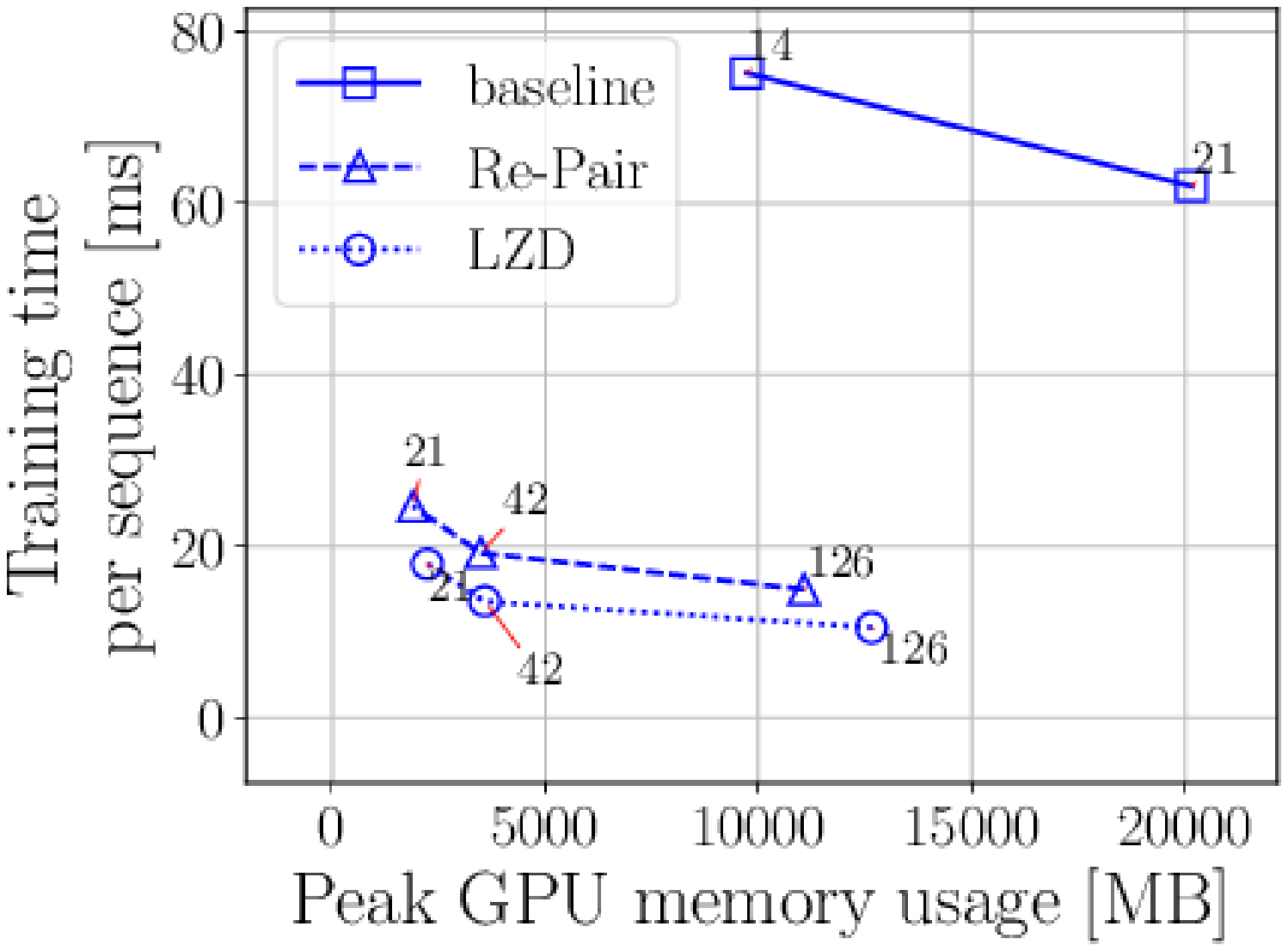}
        \subcaption{Sogou dataset}
    \end{minipage}
    \begin{minipage}[b]{0.499\linewidth}
        \centering
        \includegraphics[width=0.8\columnwidth]{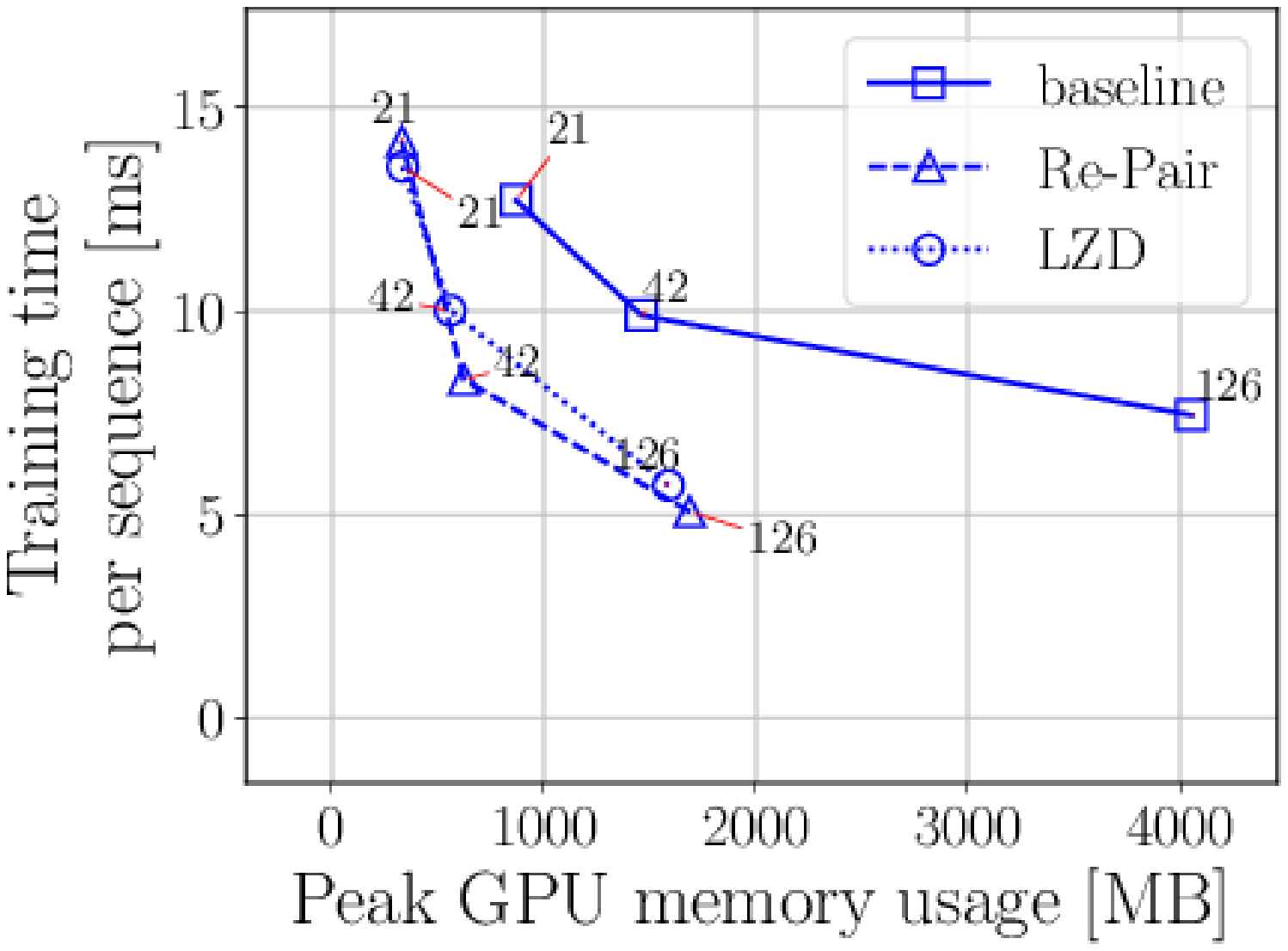}
        \subcaption{Yelp.P dataset}
    \end{minipage}
    \begin{minipage}[b]{0.499\linewidth}
        \centering
        \includegraphics[width=0.8\columnwidth]{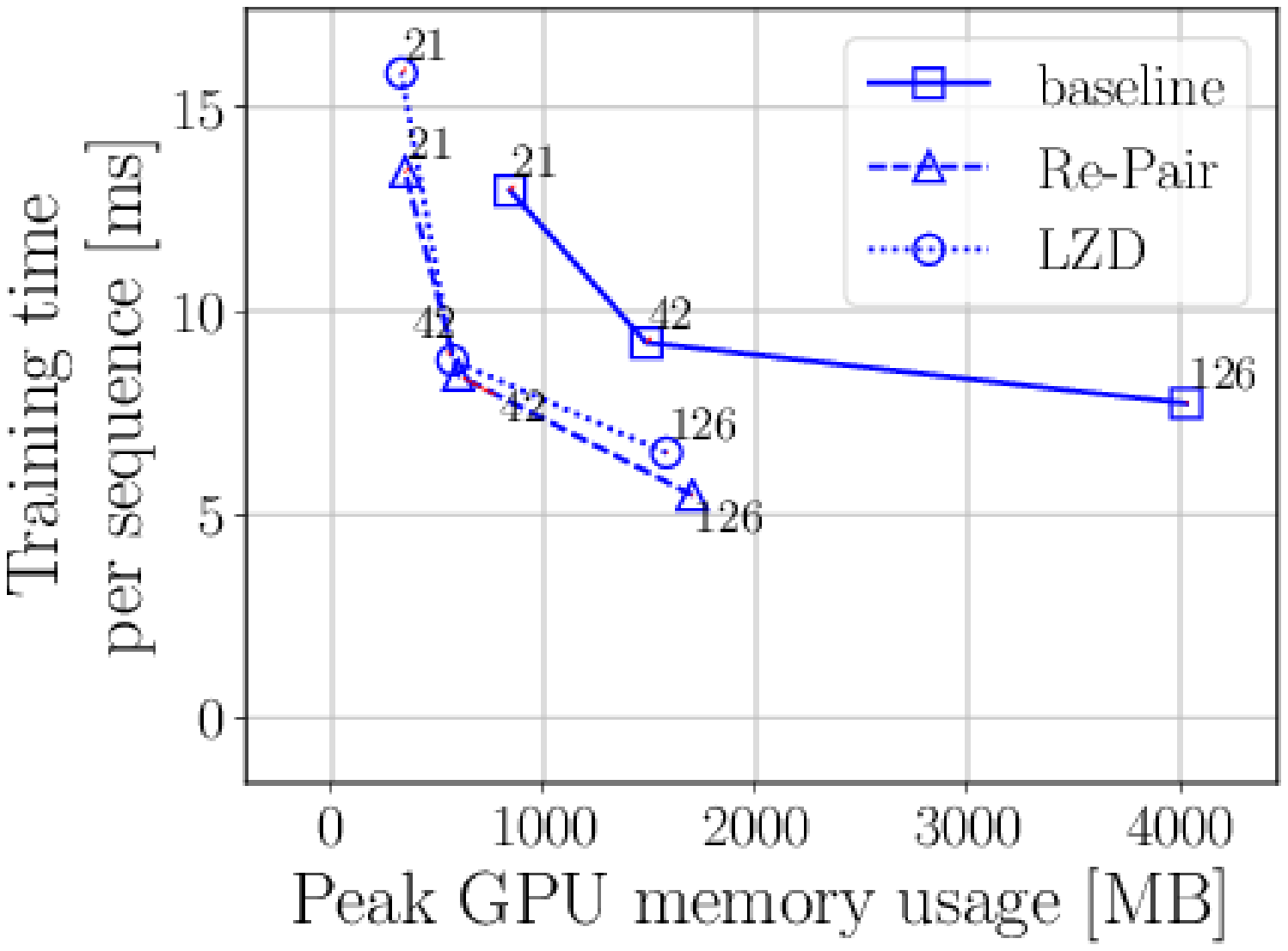}
        \subcaption{Yelp.F dataset}
    \end{minipage}
    \begin{minipage}[b]{0.499\linewidth}
        \centering
        \includegraphics[width=0.8\columnwidth]{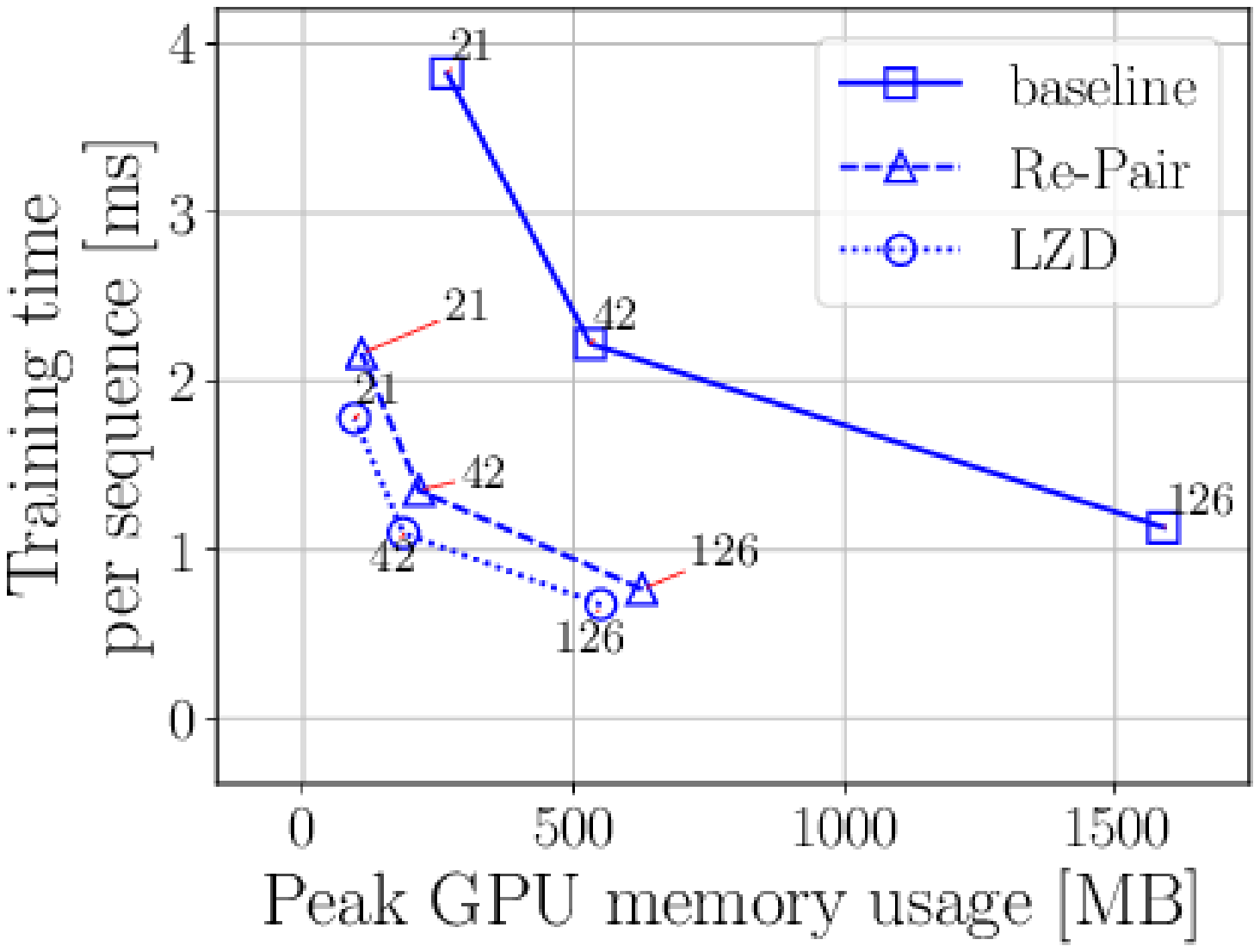}
        \subcaption{H3 dataset}
    \end{minipage}
    
    \caption{Memory usage and training time for each mini-batch size. Solid (rectangle plot) , dashed (triangle plot), and dotted line (circle plot) represent Bi-LSTM, Bi-LSTM-gru (Re-Pair), and Bi-LSTM-gru (LZD) respectively. Corresponding mini-batch size $bs$ is written beside each plot. These figures are visualized using adjustText library (\url{https://github.com/Phlya/adjustText}) to adjust label placement.
    }
    \label{fig:memory_time_figure}
\end{figure*}

\begin{table*}
    \caption{Testing accuracy (\%) in DNA classification datasets (mean $\pm$ std. dev.). $\dag$ denotes that Dual-GRU composer performs better than Naive MLP composer on the specific dataset and compression algorithm ($p<0.02$). $\ddag$ denotes that decline of performance of the best performing setting of the proposed method compared to the baseline Bi-LSTM is less than 5 points ($p<0.03$). (bracket) and \textbf{bold} denote the best performing setting of the proposed method on the specific dataset ($p>0.05$ and $p<0.05$ respectively).}
    \label{tab:performance_dna_classification}
    \centering
    \subcaption{H3, H4, H3K9ac, H3K14ac, H4ac dataset}
    \scalebox{0.8}{
        \begin{tabular}{c||c|c|c|c|c}
            Method & $^\ddag$H3 & $^\ddag$H4 & $^\ddag$H3K9ac & H3K14ac & H4ac  \\ \hline\hline
            seq-CNN & 88.99 & 88.09 & 78.84 & 78.09 & 77.40   \\ \hline\hline
            Bi-LSTM (baseline) & 84.55 $\pm$ 0.30 & 86.85 $\pm$ 0.21 & 74.60 $\pm$ 0.87 & 74.23 $\pm$ 0.34 & 71.41 $\pm$ 0.79 \\ \hline\hline
            Bi-LSTM-mlp (Re-Pair) & 78.82 $\pm$ 0.30 & 79.42 $\pm$ 0.77 & 70.17 $\pm$ 0.55 & 57.11  $\pm$ 0.43 & 56.89 $\pm$ 2.75  \\ \hline
            Bi-LSTM-gru (Re-Pair) & $^\dag$ 80.55 $\pm$ 0.27 & $^\dag$\textbf{84.50} $\pm$ 0.53 & $^\dag$\textbf{71.61} $\pm$ 0.42 & $^\dag$(64.93) $\pm$ 0.18 & $^\dag$61.61 $\pm$ 0.74  \\ \hline
            Bi-LSTM-mlp (LZD) & 80.27 $\pm$ 0.93 & 78.05 $\pm$ 0.83 & 70.55 $\pm$ 0.29 & 58.37 $\pm$ 1.60 & 59.03 $\pm$ 2.37  \\ \hline
            Bi-LSTM-gru (LZD) & (81.49) $\pm$ 0.27 & $^\dag$82.86 $\pm$ 0.17 & 70.66 $\pm$ 0.25 & $^\dag$64.29 $\pm$ 0.56 & $^\dag$(62.73) $\pm$ 0.69  \\
        \end{tabular}
    }
    \vspace{2em}
    \subcaption{H3K4me1, H3K4me2, H3K4me3, H3K36me3, H3K79me3 dataset}
    \scalebox{0.8}{
        \begin{tabular}{c||c|c|c|c|c}
            Method &  $^\ddag$H3K4me1 & $^\ddag$H3K4me2 & H3K4me3 & H3K36me3 & $^\ddag$H3K79me3 \\ \hline\hline
            seq-CNN &  74.20 & 71.50 & 74.69 & 79.26 & 83.00  \\ \hline\hline
            Bi-LSTM (baseline) &  68.03 $\pm$ 0.23 & 67.80 $\pm$ 0.71  & 64.70 $\pm$ 2.01 & 76.18 $\pm$ 0.28 & 79.21 $\pm$ 0.37  \\ \hline\hline
            Bi-LSTM-mlp (Re-Pair) &  55.18 $\pm$ 0.29 & 61.93 $\pm$ 1.71 & 53.36 $\pm$ 0.06 & 59.70 $\pm$ 2.40 & 71.29 $\pm$ 0.56 \\ \hline
            Bi-LSTM-gru (Re-Pair) &  $^\dag$63.84 $\pm$ 0.61 & 63.59 $\pm$ 0.60 & $^\dag$(58.34) $\pm$ 0.80 & $^\dag$\textbf{67.32} $\pm$ 0.58 & $^\dag$\textbf{74.87} $\pm$ 0.26 \\ \hline
            Bi-LSTM-mlp (LZD) &  55.28 $\pm$ 0.15 & 61.36 $\pm$ 2.78 & 53.28 $\pm$ 0.07 & 54.46 $\pm$ 0.00 & 60.38 $\pm$ 6.36 \\ \hline
            Bi-LSTM-gru (LZD) &  $^\dag$(64.78) $\pm$ 0.48 & $^\dag$\textbf{64.66} $\pm$ 0.40 & $^\dag$58.12 $\pm$ 1.46 & $^\dag$66.14 $\pm$ 0.36 & $^\dag$73.99 $\pm$ 0.42 \\
        \end{tabular}
    }
\end{table*}

\begin{table*}[]
    \caption{Testing errors (\%) in text classification datasets (mean $\pm$ std. dev.). $\dag$ denotes that Dual-GRU composer performs better than Naive MLP composer on the specific dataset and compression algorithm ($p<0.02$). $\ddag$ denotes that decline of performance of the best performing setting of the proposed method compared to the baseline Bi-LSTM is less than 3 points ($p<0.02$). (bracket) and \textbf{bold} denote the best performing setting of the proposed method on the specific dataset ($p>0.05$ and $p<0.05$ respectively).}
    \label{tab:performance_text_classification}
    \centering
    \scalebox{0.8}{
        \begin{tabular}{c||c|c|c}
            Method & $^\ddag$Sogou & $^\ddag$Yelp P. & Yelp F.  \\ \hline\hline
            Char-CNN (large) & 4.88 & 5.89 & 39.62 \\ \hline
            Char-CNN (small) & 8.65 & 6.53 & 40.84 \\ \hline\hline
            Bi-LSTM (baseline) & 4.18 $\pm$ 0.05  & 6.21 $\pm$ 0.12 & 39.14 $\pm$ 0.23  \\ \hline\hline
            Bi-LSTM-mlp (Re-Pair) & 6.10 $\pm$ 0.13 & 10.03 $\pm$ 0.11  & 45.63 $\pm$ 0.18 \\ \hline
            Bi-LSTM-gru (Re-Pair) & $^\dag$\textbf{5.14} $\pm$ 0.04 & $^\dag$\textbf{8.83} $\pm$ 0.11 & $^\dag$(43.98) $\pm$ 0.27 \\ \hline
            Bi-LSTM-mlp (LZD) & 6.77 $\pm$ 0.13 & 11.33 $\pm$ 0.24 & 47.21 $\pm$ 0.35  \\ \hline
            Bi-LSTM-gru (LZD) & $^\dag$5.77 $\pm$ 0.09 & $^\dag$9.26 $\pm$ 0.17 & $^\dag$44.25 $\pm$ 0.21 \\
        \end{tabular}
    }
\end{table*}

Table~\ref{tab:performance_dna_classification} and Table~\ref{tab:performance_text_classification} show the performance of the baseline (\textbf{Bi-LSTM (baseline)}) and the proposed method with naive MLP and Dual-GRU composer module (\textbf{Bi-LSTM-mlp}, \textbf{Bi-LSTM-gru}) in DNA and text classification datasets respectively.
Reported values are averages over four experiments, in which the initial network parameters and held-out development set were randomly sampled.
For all settings, we used the same set of four seed values to sample the held-out development set.
Also, we used Wilcoxon rank sum test \cite{wilcoxon1945individual} with multiple-test adjustment using Holm's method \cite{holm1979simple} to check the statistical significance of the results.

We include the reported performances of character-level CNN \cite{DBLP:conf/nips/ZhangZL15} (\textbf{Char-CNN}) in the text classification datasets, and those of \textbf{seq-CNN} \cite{DBLP:conf/naacl/Johnson015} in the DNA classification datasets reported in \cite{nguyen2016dna}

As shown in Table~\ref{tab:performance_text_classification} and Table~\ref{tab:performance_dna_classification}, the performance drop of the best performing setting of the proposed method compared to the baseline Bi-LSTM was significantly less than three points on two out of three text dataset; and less than five points on six out of ten DNA datasets.
Dual-GRU composer performed significantly better than naive MLP composer on all combinations of the text datasets and the compression algorithms; and on 17 out of 20 combinations of the DNA dataset and the compression algorithms, which suggests the importance of designing good composer modules.

\subsubsection*{Memory and Computational Efficiency}

We also investigated the memory and computational efficiency of the proposed method compared to the baseline Bi-LSTM on non-compressed sequences.s
Since the mini-batch size $bs$ influences the memory usage and training time, we tested three different mini-batch sizes $bs=\{ 21,42,126 \}$.\footnote{Since min-batch sizes $bs=\{ 42, 126 \}$ caused GPU memory overflow in the Sogou dataset, we conducted experiments with smaller mini-batch sizes $bs=\{ 14, 21 \}$ on the dataset.}
Setting of the other hyperparemters are kept same as the setting in the experiments in Section~\ref{subsubsec_classification_performance}.
Also, we did not include decompression time to the results of the baseline method.
Experiments of text datasets were conducted with a Tesla V100-SXM2 with a 32GB memory, and experiments of DNA datasets were conducted with Tesla V100-PCIE with 16GB memory.

Figure \ref{fig:memory_time_figure} shows peak GPU memory usage and required training time during training phase of the baseline method (Bi-LSTM) and the proposed method (Bi-LSTM-gru) with two compression algorithms (Re-Pair and LZD)\footnote{Since results of 10 DNA datasets are almost identical, we only show the result of H3 dataset. Note that statistics of sequence length and number of replacement rules of 10 DNA datasets are almost identical.}.
As shown in Figure~\ref{fig:memory_time_figure}, with a fixed size of GPU memory, compared to the baseline method, training time of the proposed method is around five times faster in Sogou, around two times faster in Yelp.P and Yelp.F, and around 3 times faster in the DNA datasets respectively.
These results agrees with complexity analysis in Section~\ref{subsec:complexity} that the proposed method is more memory and computationally efficient on the datasets with higher compression ratio (e.g. Sogou) than the datasets with moderate compression ratio (e.g. Yelp.P and Yelp.F).
\section{Conclusion}
In this paper, we proposed a method to directly apply sequence models to grammar compressed sequence data without decompression.
To encode a compressed sequence into a sequence of vector representations that sequence models can handle, the proposed method incrementally encode unique symbols in compression rules into the vector representation by means of a composer module.

We empirically showed that, by applying the proposed method to compressed sequences, we can reduce the required training time and GPU memory compared to sequence models applied to original decompressed sequences. 

There are several possible directions for future work.
First, we will consider applying sequence models for compressed text using another kind of compression technique such as zip encoding, which is not grammar compression.
Finally, we will investigate the performance of the proposed method with shared replacement rules across a whole dataset.

\subsection*{Acknowledgement}
The authors are grateful to Keigo Kimura, Carolin Lawrence, Yuzuru Okajima, and Kunihiko Sadamasa for their comments and discussion, which significantly improved the correctness and the presentation of this paper.
\newline
\bibliographystyle{main}
\bibliography{main}

\end{document}